\documentclass{article} 
\usepackage{iclr2027_conference,times}

\usepackage{amsmath,amsfonts,bm}

\def\eqref#1{equation~\ref{#1}}

\def\1{\bm{1}}

\DeclareMathAlphabet{\mathsfit}{\encodingdefault}{\sfdefault}{m}{sl}
\SetMathAlphabet{\mathsfit}{bold}{\encodingdefault}{\sfdefault}{bx}{n}

\usepackage{hyperref}
\usepackage{url}
\usepackage{microtype}
\usepackage{graphicx}
\usepackage{wrapfig}
\usepackage{float}
\usepackage{booktabs}
\usepackage{multirow}
\usepackage{array}
\usepackage{amsmath,amssymb,amsthm}
\usepackage{xcolor}
\usepackage{placeins}
\usepackage{pifont}

\title{Learning from the Gap Between Pass@K and Pass@1}

\author{
  Xuan Liu\\
  Shanghai Jiao Tong University\\
  \texttt{liuxuan\_cn@outlook.com}
  \And
  Jingbin Qian\\
  William Marsh Rice University\\
  \texttt{jingbinqian2002@gmail.com}
  \And
  Haosheng Chen\\
  East China Normal University\\
  \texttt{hschen@stu.ecnu.edu.cn}
}

\iclrfinalcopy
\makeatletter
\long\def\@makecaption#1#2{%
  \vskip\abovecaptionskip
  \sbox\@tempboxa{#1: #2\strut}%
  \ifdim \wd\@tempboxa >\hsize
    #1: #2\strut\par
  \else
    \global \@minipagefalse
    \hb@xt@\hsize{\hfil\box\@tempboxa\hfil}%
  \fi
  \vskip\belowcaptionskip}
\makeatother

\begin{document}

\maketitle
\lhead{Preprint}
\begingroup
\renewcommand{\thefootnote}{}
\footnotetext{Code: \url{https://github.com/Xuanxuana1/GapFT}}
\endgroup

\begin{abstract}
Sampling many responses and keeping one that passes a verifier lets
large language models solve problems beyond their single-response ability,
but this search must be paid again for every query, while many deployments
answer with a single response.  Post-training on verified responses can
transfer the benefit of search into the model. With a fixed budget, selecting by
correctness alone spends slots on problems the model already answers correctly,
leaving fewer to correct its failures. To address this imbalance, we propose
GapFT, which
\textbf{trains on the gap between Pass@$K$ and Pass@1}: problems
that the source model fails with one response but solves within $K$ samples.
GapFT keeps the objective and training budget fixed and changes only which
verified responses enter training; an exact decomposition splits the
resulting Pass@1 change into corrected failures and regressions on problems
the source model already solved.  On LogiQA~2.0 and ReClor with three
model families, GapFT is above budget-matched uniform rejection-sampling
fine-tuning (RFT) in every setting, with a positive pooled effect, and on
Llama-3.1-8B and Mistral-7B it recovers about two thirds to four fifths of the
gain of fine-tuning on the entire verified pool with 11--34\% of its problems.
Further analyses reveal that the gain comes from failures that the first few
search samples recover, while failures found only by deeper search displace
replay and add no net gain, that filling the same budget with gold-labeled
failures search cannot reach lowers accuracy, and that the gain is
bounded by how many transferable failures search exposes.
\end{abstract}

\section{Introduction}
\label{sec:introduction}

Large language models (LLMs) now solve many reasoning problems in
mathematics, logic, and code, yet a single response often fails on problems
the model can still solve: when several responses are sampled and a
verifier checks each one, a correct response frequently appears among them
\citep{chen2021codex,li2022alphacode}.  This gap between Pass@$K$ and Pass@1
has made additional inference computation a standard way to improve
accuracy, from aggregating independent paths with self-consistency to scaling
the number of samples and verifications at test time
\citep{wang2022selfconsistency,snell2024testtime}.  Search can also branch
through changing environments, as in tool-using agents \citep{yao2023react},
or keep the input fixed and explore alternative responses that a
verifier can check.  We study the latter setting.  For a problem $x$, every sampled response receives
the same $x$; the source model is queried once for the deployment answer and
$K$ times for search, with no feedback between generations.

\begin{figure}[t]
\centering
\includegraphics[width=\linewidth]{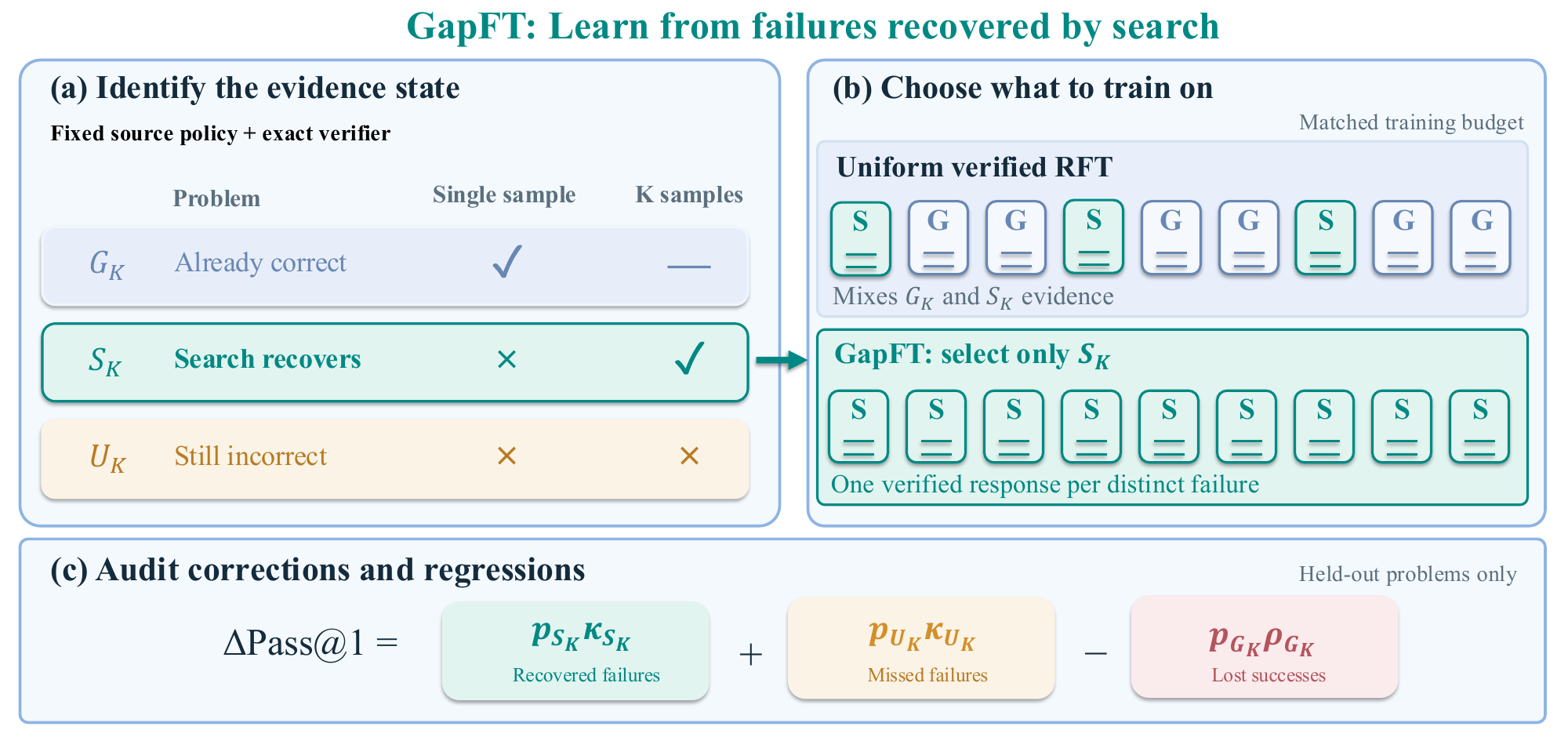}
\caption{GapFT selects verified evidence for single-sample decoding. (a) The verifier checks one direct response and $K$ sampled responses from the source model, placing each problem in $G_K$, $S_K$, or $U_K$; (b) under a matched training budget, GapFT selects one verified response for each distinct $S_K$ problem, whereas uniform RFT mixes $G_K$ and $S_K$ evidence; (c) the held-out Pass@1 change decomposes into corrections and regressions.}
\label{fig:overview}
\end{figure}

The accuracy gained by search, however, must be paid for again at every query,
while many
deployments answer with a single response.  Post-training can amortize the
cost by searching a training candidate pool once and using its verified
responses to update the policy, so that later problems need less search.  Offline
methods such as STaR and ReST retain successful model-generated traces
\citep{zelikman2022star,gulcehre2023rest}, while RLVR methods such as ReFT and
GRPO use verifier feedback during optimization
\citep{trung2024reft,shao2024deepseekmath}.
Other work distills a selected Best-of-$N$ distribution or studies the value of
distinct correct paths \citep{sessa2025bond,yuan2023scaling,amini2025vbon}.
Together, these methods establish that search-derived correct responses can
improve a policy.  They leave open a more basic data question: when the
training budget is fixed, which verified responses should occupy its limited
slots?

These recipes judge a verified response by properties of the response
itself: whether it is correct, how much reward or preference it receives, or
how much it differs from other correct responses
\citep{zelikman2022star,gulcehre2023rest,sessa2025bond,yuan2023scaling}.  Yet
the purpose of amortization is to change what the model does in a single
attempt, and \textbf{whether a verified response can change single-attempt
behavior depends on the source model, not on the response alone.}  On a
problem the source model already answers correctly, a verified response
repeats behavior the model already has; on a problem it misses in one attempt
but solves by search, the same kind of response supplies exactly the target
that single-response deployment lacks.  Existing selection criteria do not
separate the two cases, and because problems already solved in one attempt
usually outnumber those recovered only by search, allocation without this
distinction spends much of a fixed training budget on redundant examples.  What is
missing is therefore not another way to obtain correct responses, but a
principle for allocating them relative to what the deployed model already
does; we call this the data-selection gap.

To address this data-selection gap, we design GapFT to condition evidence
selection on the source policy's single-sample outcome.  Specifically, we use the verifier to
check one direct response and $K$ sampled responses from the source model on
each candidate problem, and place the problem in $G_K$ (single-sample
success), $S_K$ (single-sample failure recovered by search), or $U_K$ (failure
missed by the search budget; Figure~\ref{fig:overview}).  GapFT then fills
the post-training budget with verified responses from $S_K$ problems, leaving
the training objective unchanged.

Empirically, we systematically study this allocation on LogiQA~2.0 and ReClor
with Llama-3.1-8B, Qwen2.5-7B, and Mistral-7B.  All conditions
share the same training objective and budget, so differences in
single-response accuracy reflect only which verified responses enter
training.  To locate where each change comes from, we decompose it exactly
into corrections of failures that search recovers, corrections of failures
that search misses, and regressions on problems the source model already
solves.  In summary, our contributions are as follows:
\begin{itemize}
    \item We make the source checkpoint's single-sample outcome an explicit variable in
    verified-data allocation.  This exposes the redundancy between $G_K$ and
    $S_K$ evidence that fixed-budget recipes otherwise mix.
    \item We introduce GapFT, a matched-budget recipe that selects verified
    responses from $S_K$ while keeping the objective fixed, together with an
    exact decomposition of correction and regression and controls for search
    reachability.
    \item On two logic benchmarks and three model families, GapFT is above
    uniform rejection-sampling fine-tuning (RFT) at a matched budget and
    recovers most of the full verified pool's gain with a fraction of its
    problems.  We further show that gains come from covering distinct
    reachable failures and are bounded by the transferable failures search
    exposes.
\end{itemize}

\section{Related Work}
\label{sec:related_work}

\paragraph{Search and verification at inference time.}
Pass@$K$ made repeated sampling an explicit source of performance
\citep{chen2021codex}; self-consistency, Tree of Thoughts, and AlphaCode
aggregate, branch, or filter many samples
\citep{wang2022selfconsistency,yao2023tree,li2022alphacode}.  Outcome verifiers
score solutions \mbox{\citep{cobbe2021gsm8k,hosseini2024vstar}}; process
verifiers score steps \mbox{\citep{lightman2023verify}}.  Recent work
allocates inference compute by difficulty and measures the return from
verification \citep{snell2024testtime,setlur2025verification}.  These methods
spend compute at every query.  GapFT runs the same search once, on
candidate problems, and asks what its verified output is worth after search is
removed from deployment.

\paragraph{What verified post-training trains on.}
STaR, ReST, and rejection-sampling fine-tuning train on responses a verifier
accepts \citep{zelikman2022star,gulcehre2023rest,shao2024deepseekmath}.
ReFT uses online PPO with answer-based rewards rather than simply retaining
accepted offline responses \citep{trung2024reft}.
\citet{yuan2023scaling} observe that the gain tracks the number of distinct
correct paths rather than the number of samples.  DART-Math allocates more
synthesis trials to difficult queries so that rejection-sampled data are not
dominated by easy ones \citep{tong2024dartmath}.  Online methods filter
implicitly: DAPO drops prompts whose sampled group is entirely correct or
incorrect \citep{yu2025dapo}, and prompt curricula choose which problems
receive updates \citep{gao2025pcl}.  Distillation imitates a Best-of-$N$
distribution or a specified selection rule
\citep{sessa2025bond,amini2025vbon,chow2025inferenceaware}, SCoRe shows that
correction traces collected in advance can mismatch the deployed policy
\citep{kumar2024score}, and Learning From Failure trains on unsuccessful
trajectories \citep{wang2024learningfailure}.  These methods establish the
general strategy of transferring search-derived evidence into the policy, but
none holds training examples, processed tokens, and optimizer steps fixed while
measuring the Pass@1 value of a verified response as a function of whether the
deployed policy already produces it.  This is the quantity GapFT isolates
(Appendix Table~\ref{tab:related_comparison}).

\paragraph{Amplification, interference, and regression.}
Whether post-training amplifies behavior already present or acquires new
behavior is debated: large-$K$ comparisons and Echo Chamber report
amplification \citep{yue2025limitrlvr,zhao2025echo}, whereas synthetic
experiments show unseen compositions \citep{yuan2025composition}.
\citet{2026passk_degrades_pass1} show that Pass@$k$ objectives can degrade
Pass@1 through prompt interference, ReGenesis links weak transfer to
task-specific self-generated reasoning \citep{peng2024regenesis}, and the
alignment-tax and forgetting literature quantifies what adaptation costs
\citep{lin2023alignmenttax,kalajdzievski2024forgetting}.
Our decomposition places these quantities in one identity measured on held-out
items: amplification is correction on $S_K$, acquisition beyond observed
support is correction on $U_K$, and the cost is regression on $G_K$.

\section{Methodology: GapFT}
\label{sec:method}

Section~\ref{sec:preliminaries} fixes notation for the offline rejection-
sampling fine-tuning (RFT) setting used by GapFT
\citep{zelikman2022star,gulcehre2023rest}.  GapFT then adds two
parts: an intervention that decides which verified evidence fills a fixed post-training
budget, and a decomposition of the resulting Pass@1 change for each audit item
(Figure~\ref{fig:overview}).  The training objective is unchanged.

\subsection{Problem Setup}
\label{sec:preliminaries}

Let $V(x,y)\in\{0,1\}$ be an exact outcome verifier for problem $x$ and
response $y$, and let $d_\pi(x)$ be the response of policy $\pi$ under the
single-sample decoding (a greedy answer for logic and MATH~\citep{hendrycks2021math}, the final
submission of one bounded episode for code).  Under the fixed-input search
setting, repeated sampling draws $\mathcal{Y}^{\pi}_K(x)=(y_1,\ldots,y_K)$ as
conditionally independent samples from $\pi(\cdot\mid x)$: every generation
receives the same $x$ and no feedback from the other generations.  We count
$x$ as solved if any sample passes \citep{chen2021codex}.  For a problem
distribution $\mathcal{D}$, deployment accuracy and search success are
\begin{equation}
A_{\mathrm{dep}}(\pi)=\mathbb{E}_{x\sim\mathcal{D}}[V(x,d_\pi(x))],\qquad
A_{\mathrm{search}}(\pi;K)=\mathbb{E}_{x\sim\mathcal{D},\,y_{1:K}\sim\pi(\cdot\mid x)}
\Big[\max_{k\leq K} V(x,y_k)\Big],
\label{eq:pass_metrics}
\end{equation}
Pass@1 and Pass@$K$ in the usual terminology, and their difference is the
Pass@$K$--Pass@1 gap.

The offline RFT procedure studied here samples responses from a frozen source
policy $\pi_{\mathrm{src}}$ on a training candidate pool $\mathcal{D}_{\mathrm{cand}}$,
keeps the pairs accepted by the verifier as a training set $\mathcal{T}$, and
fine-tunes with the maximum-likelihood objective
\begin{equation}
\mathcal{L}(\theta)
=-\sum_{(x,y)\in\mathcal{T}}\sum_{t=1}^{|y|}
  \log \pi_\theta(y_t\mid x,y_{<t})
\label{eq:rft_objective}
\end{equation}
\citep{zelikman2022star,gulcehre2023rest,yuan2023scaling,shao2024deepseekmath}.
Standard RFT admits a response on correctness alone and, when the budget is
smaller than the accepted pool, subsamples regardless of whether
$\pi_{\mathrm{src}}$ already solves the problem in one decode.  We write $\pi_{\mathcal{T}}$ for the policy trained on $\mathcal{T}$ and evaluate
it on an audit set $\mathcal{D}_{\mathrm{audit}}$ disjoint from
$\mathcal{D}_{\mathrm{cand}}$.

\subsection{Evidence States}
\label{sec:evidence_states}

Relative to the source policy and a search budget $K\geq1$, we place every
problem in exactly one of three \emph{evidence states},
\begin{align}
G_K &= \{x: V(x,d_{\mathrm{src}}(x))=1\}, \\
S_K &= \{x: V(x,d_{\mathrm{src}}(x))=0,\
              \exists y\in\mathcal{Y}^{\pi_{\mathrm{src}}}_K(x):V(x,y)=1\},\\
U_K &= \{x: V(x,d_{\mathrm{src}}(x))=0,\
              \forall y\in\mathcal{Y}^{\pi_{\mathrm{src}}}_K(x):V(x,y)=0\},
\end{align}
the \emph{deployed successes}, the \emph{failures recovered by search}, and
the \emph{failures missed by search}.  The partition records observations
under a finite search budget, not capability labels, so $U_K$ does not mean
the source assigns zero probability to a correct response.  A verified
response on a deployed success is redundant for deployment; one on a failure
recovered by search is a target the deployed policy does not yet produce.
GapFT rests on the hypothesis that this distinction, not correctness
alone, determines training value.

\subsection{Evidence Selection under a Matched Budget}
\label{sec:gapft}

Search runs only on $\mathcal{D}_{\mathrm{cand}}$ and yields two evidence sets,
\begin{equation}
\mathcal{E}_G=\{(x,d_{\mathrm{src}}(x)):x\in\mathcal{D}_{\mathrm{cand}}\cap G_K\},\qquad
\mathcal{E}_{S,K}=\{(x,y_K^*(x)):x\in\mathcal{D}_{\mathrm{cand}}\cap S_K\},
\label{eq:evidence_sets}
\end{equation}
where $y_K^*(x)$ is one verified response chosen by a rule frozen before
training (the shortest verified response for logic and code; a frozen
generation order for MATH).

Let $n=|\mathcal{T}|$ be the number of distinct examples in a training set and
$R$ the number of times it is repeated.  We describe the post-training budget
by $\mathcal{B}=(N,L,J)$: $N=Rn$ example exposures, $L$ processed tokens, and
$J$ optimizer steps; the search budget $K$ is reported separately.  GapFT sets
$\mathcal{T}_{\mathrm{gap}}=\mathcal{E}_{S,K}$, so each recovered failure
contributes one verified response.  Two baselines draw the same $n$ examples
without replacement from the same pool.  \emph{Uniform RFT} samples
$\mathcal{E}_G\cup\mathcal{E}_{S,K}$ regardless of state, as standard RFT does
whenever the budget is smaller than the pool; \emph{solved replay}, the
no-search condition $K=0$, samples $\mathcal{E}_G$ only.  Across conditions we
enforce
\begin{equation}
N=R|\mathcal{T}|,\qquad
R\operatorname{Tok}(\mathcal{T})\simeq L,\qquad
\operatorname{Step}(\mathcal{T};R)=J,
\label{eq:matched_budget}
\end{equation}
where $\operatorname{Tok}$ counts input and response tokens after truncation
and $\operatorname{Step}(\mathcal{T};R)$ counts updates over $R$ repetitions,
so that conditions differ only in which evidence fills the budget.
\emph{RFT on all verified items} trains on $\mathcal{E}_G\cup\mathcal{E}_{S,K}$
with the same $R$ and serves as a larger-budget reference.  All conditions
minimize Equation~\ref{eq:rft_objective}, and GapFT's effect against a
baseline $\mathcal{T}_b$ is
$\Delta_b=A_{\mathrm{dep}}(\pi_{\mathcal{T}_{\mathrm{gap}}})-A_{\mathrm{dep}}(\pi_{\mathcal{T}_b})$.
Audit responses never enter evidence selection, training, or checkpoint
selection.

\subsection{Decomposition of the Pass@1 Change}
\label{sec:decomposition}

We introduce an identity that attributes any Pass@1 change to the three
states.  On the audit set, partition problems into $G_K$, $S_K$, and $U_K$
using the source greedy decision and $K$ source samples, and let
$p_B=\Pr_{x\sim\mathcal{D}_{\mathrm{audit}}}[x\in B_K]$.  For a post-training
policy $\pi$ define
\begin{equation}
\kappa_B=\Pr[V(x,d_\pi(x))=1\mid x\in B_K]\ \text{for }B\in\{S,U\},\qquad
\rho_G=\Pr[V(x,d_\pi(x))=0\mid x\in G_K],
\label{eq:rates}
\end{equation}
the correction rates on the two failure states and the regression rate on
deployed successes.  Because audit items are disjoint from training,
$\kappa_S$ measures generalization rather than reproduction.  Since the three
states partition the audit set,
\begin{equation}
A_{\mathrm{dep}}(\pi)-A_{\mathrm{dep}}(\pi_{\mathrm{src}})
=p_S\kappa_S+p_U\kappa_U-p_G\rho_G
\label{eq:decomposition}
\end{equation}
holds exactly, and any paired contrast $\Delta_b$ decomposes into the
differences of the three terms.  The weights bound each term; in particular
$p_S\kappa_S\leq p_S$, so no post-training can gain more from correction
within observed support than the share of failures recovered by search in the
audit set.  The identity explains a measured contrast but does not assign a
cause; causality comes from the matched-budget interventions of
Section~\ref{sec:mechanism} (Appendix Figure~\ref{fig:audit_flow}).

\noindent\textbf{Remark.} Three lines of work are close to GapFT but act at
different levels.  \citet{2026passk_degrades_pass1} show that optimizing
Pass@$K$ can lower Pass@1 through gradient conflict between prompts;
\citet{yue2025limitrlvr} find that RL mainly sharpens correct paths the base
model already supports; DART-Math \citep{tong2024dartmath} spends more
synthesis trials on queries that a separate model often fails.  GapFT keeps
the objective, the pool, and the budget $(N,L,J)$ fixed and decides only which
problems fill the budget.  Its signal is not difficulty but whether the
trained policy fails in one decode yet reaches the answer by search.
Section~\ref{sec:mechanism} suggests the distinction matters: filling a small
budget with failures chosen regardless of reachability lowers Pass@1.

\section{Experiments}
\label{sec:main_results}

\subsection{Setup}
\label{sec:exp_setup}

\paragraph{Tasks and models.}
We use LogiQA~2.0 \citep{liu2023logiqa2} and ReClor \citep{yu2020reclor},
two multiple-choice logical reasoning benchmarks whose answers are verified
exactly by matching the chosen option.  Search runs on the training split, and
all conditions are evaluated on a disjoint audit split: the 500 ReClor
validation problems and the 1{,}367 LogiQA~2.0 test problems whose prompt never
appears in its training split.  We also exclude the 211 LogiQA training
records whose prompts appear in the original test split from the training
candidate pool.  The source policies are
Llama-3.1-8B-Instruct \citep{grattafiori2024llama3}, Qwen2.5-7B-Instruct
\citep{qwen2024qwen25}, and Mistral-7B-Instruct-v0.3 \citep{mistral2024v03}.

\paragraph{Search and training.}
Following the repeated-sampling protocol of Pass@$K$ \citep{chen2021codex},
the source decodes each candidate problem greedily once and samples $K=8$
responses; the shortest verified response, a rule fixed before training, is
retained.  Following offline RFT
\citep{yuan2023scaling,shao2024deepseekmath}, each condition fine-tunes the
source with LoRA \citep{hu2022lora} for one epoch over its training set
repeated $R=8$ times, using the same learning rate $5{\times}10^{-5}$ without
per-condition tuning and three training seeds.

\paragraph{Budget and evaluation.}
The main comparison sets $n=|\mathcal{E}_{S,8}|$, so GapFT, uniform RFT, and
solved replay share $N$ and $J$ exactly (Appendix Table~\ref{tab:logic_budgets}).
Recovered failures fill about a third of each Llama pool, an eighth of each
Mistral pool, and under 4\% of each Qwen pool, which is the share uniform RFT
draws by chance.  We report greedy Pass@1 on the audit split with 95\% intervals
from 2{,}000 bootstrap replicates \citep{efron1994bootstrap} that resample
training seeds and paired audit problems.
Appendix~\ref{app:implementation} lists the full settings.

\subsection{Main Results}
\label{sec:exp_main}
\label{sec:amortization}

\begin{table*}[t]
\centering
\footnotesize
\setlength{\tabcolsep}{2pt}
\caption{Pass@1 Under Matched Evidence Budgets. Among Source, Replay,
Uniform, and GapFT, the best result in each row is \textbf{bold} and the
second-best is \underline{underlined}. All is a larger-budget reference
excluded from this ranking. Brackets give 95\% confidence intervals.}
\label{tab:main}
\begin{tabular*}{\textwidth}{@{\extracolsep{\fill}}llrrrrrrrl}
\toprule
& & \multicolumn{2}{c}{Problems} & \multicolumn{5}{c}{Pass@1 (\%)} & GapFT$-$Source \\
\cmidrule(lr){3-4}\cmidrule(lr){5-9}
Model & Task & $n$ & All & Source & Replay & Uniform & GapFT & All & (pp) \\
\midrule
Llama-3.1-8B & LogiQA 2.0 & 3{,}190 & 9{,}429 & 50.33 & 55.84 & \underline{60.33} & \textbf{62.59} & 65.89 & $+12.27$ [$10.36$,\,$14.24$] \\
& ReClor & 913 & 3{,}050 & 55.40 & 61.27 & \underline{69.13} & \textbf{70.67} & 73.80 & $+15.27$ [$12.13$,\,$18.33$] \\
Qwen2.5-7B & LogiQA 2.0 & 275 & 8{,}797 & \underline{68.54} & 68.50 & 68.20 & \textbf{68.81} & 70.06 & $+0.27$ [$-0.59$,\,$1.12$] \\
& ReClor & 63 & 2{,}977 & \underline{72.20} & 72.00 & 71.53 & \textbf{72.47} & 73.20 & $+0.27$ [$-0.73$,\,$1.33$] \\
Mistral-7B & LogiQA 2.0 & 887 & 6{,}554 & 47.26 & \underline{49.74} & 49.57 & \textbf{51.87} & 53.21 & $+4.61$ [$3.27$,\,$6.00$] \\
& ReClor & 252 & 2{,}192 & 48.80 & \underline{53.47} & 52.53 & \textbf{54.33} & 57.20 & $+5.53$ [$3.60$,\,$7.53$] \\
\midrule
\multicolumn{9}{l}{Descriptive equal-weight mean of GapFT$-$Source over the six cells} & $+6.37$ \\
\bottomrule
\end{tabular*}
\end{table*}

\begin{wrapfigure}{R}{2.4in}
  \vspace{-0.35cm}
  \centering
  \includegraphics[width=2.3in]{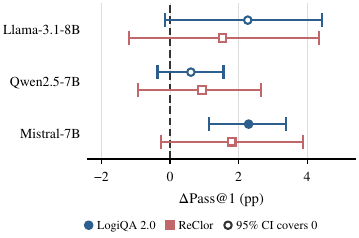}
  \vspace{-0.25cm}
  \caption{\textbf{GapFT above uniform RFT at a matched budget.} Each point
  is the Pass@1 difference between GapFT and uniform RFT, with a 95\%
  bootstrap interval over seeds and audit problems; filled markers
  have intervals that exclude zero.}
  \label{fig:main_result}
  \vspace{-0.6cm}
\end{wrapfigure}

\paragraph{Filling the budget with recovered failures raises single-sample accuracy.}
At the same number of problems, exposures, and updates, GapFT is above uniform
RFT on both tasks for all three models (Table~\ref{tab:main},
Figure~\ref{fig:main_result}).  With three seeds, the GapFT--uniform intervals
in Figure~\ref{fig:main_result} are wide; only Mistral--LogiQA excludes zero.
Their equal-weight mean, a descriptive summary added after the fact, is
$+1.57$ [$0.77$, $2.45$] points.  Solved replay shows where the
advantage comes from.  Uniform RFT beats replay only for Llama, whose pools are
a third recovered failures; for Mistral and Qwen a uniform draw contains few of
them and is no better than replay.  A training slot is thus worth little unless
it holds a problem the source still fails, which is exactly the rule GapFT uses.

\paragraph{Corrections outweigh the regression they bring.}
The decomposition explains the ordering (Appendix
Table~\ref{tab:main_decomposition}).  Relative to uniform RFT, GapFT corrects
more source failures, including audit failures that search itself missed, so
what it learns is not tied to the problems search recovered.  On Llama it also
regresses more on deployed successes, plausibly because none of its examples
rehearses behavior the source already has, but the added corrections are
larger.  Mistral gains entirely from correction, with no added regression.
Appendix~\ref{app:before_after} illustrates corrections and regressions
with complete before-and-after solutions on MATH.

\begin{figure*}[t]
\centering
\includegraphics[width=\textwidth]{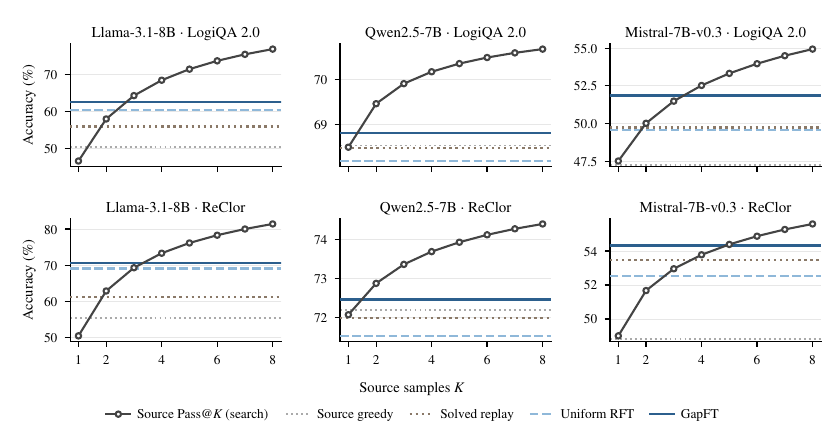}
\caption{\textbf{One tuned decode against source search.} Curves show the
source model's accuracy when a verifier picks among $K$ samples; horizontal
lines show the accuracy of one greedy decode from the source and from each
trained model.}
\label{fig:amortization}
\end{figure*}

\paragraph{The full pool buys retention with a much larger budget.}
Training on every verified problem remains the most accurate condition, at
three to almost fifty times GapFT's budget.  With 11--34\% of those problems, GapFT
recovers about two thirds to four fifths of the full pool's gain over the source on
Llama and Mistral.  On Llama, where the remaining gap is largest, the full pool
corrects recovered failures at a rate similar to GapFT and gains mainly through
lower regression, so its extra deployed successes preserve behavior rather than
add corrections.  Replaying such examples alongside GapFT is a natural use of a
larger budget; GapFT decides what to train on when the budget is small.  In
inference terms, one GapFT response on these two models reaches the source's
verifier-selected Pass@2 to Pass@4 (Figure~\ref{fig:amortization}, Appendix
Table~\ref{tab:amortization}).

\paragraph{Allocation cannot create failures to correct.}
Qwen2.5-7B is the boundary case.  Search recovers only about 2\% of its audit
problems, every condition stays within about one and a half points of the
source, and GapFT's advantage over uniform RFT is not resolved.
Section~\ref{sec:boundaries} traces this limit to how many failures search
exposes.

\section{Why GapFT Works: Covering Reachable Failures}
\label{sec:mechanism}

The controls in this section ask what makes a GapFT training slot valuable.
All three reuse the Llama recipe of Table~\ref{tab:main}, with its budget and
training seeds, and change only which problems fill the budget.

\paragraph{Control design.}
The search-depth control keeps only the failures that the first search sample
recovers, with GapFT's response for each, and fills the remaining slots with a
prefix of the solved-replay recipe; $K=0$ is solved replay and $K=8$ is GapFT.
The randomized control keeps the failures recovered within the first four
samples and includes a random $0/50/100\%$ of those first recovered in samples
five through eight, again filling omitted slots with replay; $100\%$ is GapFT.
$N$ and $J$ are unchanged, so any difference is caused by which problems are
covered.  The gold-label control replaces GapFT's
problems with the same number of source failures drawn from $S_8\cup U_8$
using gold labels, uniformly within each gold-answer class; about half of the
drawn problems are failures that search missed.  Each problem receives
GapFT's answer-only target string with the same option label, for example
\texttt{A} or \texttt{A.}, so the two recipes share their target strings, $N$,
supervised response tokens, and $J$.  Two audit checks guard against artifacts
of multiple-choice tasks: post-training answers are permuted within strata
defined by the source answer and audit group, preserving answer marginals while
breaking problem--answer pairing, and the $S_K/U_K$ split is recomputed from
four independent source sample sets with all decisions fixed.

\begin{figure*}[t]
\centering
\includegraphics[width=0.88\textwidth]{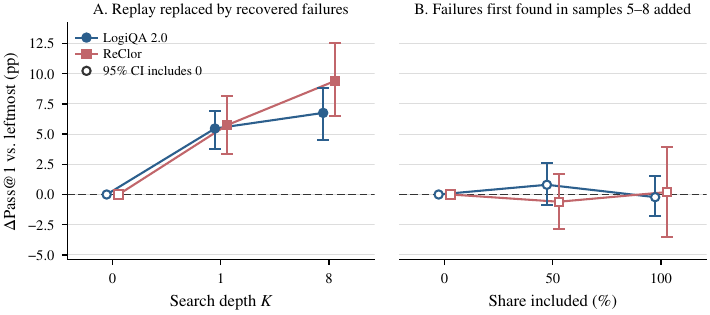}
\caption{\textbf{Accuracy as each control moves toward GapFT.} Each point is
the Pass@1 change of Llama-3.1-8B relative to the leftmost setting of its
panel, which fills the varied slots with solved replay: (A) $K=0$; (B) none
of the failures first recovered in samples 5--8. The rightmost point is
GapFT. Bars are 95\% bootstrap intervals over training seeds and audit
problems; filled markers have intervals that exclude zero.}
\label{fig:mechanism_controls}
\end{figure*}

\begin{table*}[t]
\centering
\small
\setlength{\tabcolsep}{4.0pt}
\caption{Pass@1 Decomposition of the Evidence Controls}
\label{tab:channels}
\begin{tabular*}{\textwidth}{@{\extracolsep{\fill}}llrlrrr}
\toprule
Dataset & Contrast & $\Delta$ Pass@1 (pp) & 95\% CI & $G$ & $S$ & $U$ \\
\midrule
\multicolumn{7}{l}{\textit{A. Search depth, replay replaced by recovered failures, main recipe}} \\
LogiQA 2.0 & $K{=}1-K{=}0$ & $\mathbf{+5.46}$ & [3.78, 6.95] & $-1.19$ & $+3.97$ & $+2.68$ \\
LogiQA 2.0 & $K{=}8-K{=}0$ & $\mathbf{+6.75}$ & [4.49, 8.80] & $-5.75$ & $+6.39$ & $+6.12$ \\
ReClor & $K{=}1-K{=}0$ & $\mathbf{+5.73}$ & [3.33, 8.13] & $-2.20$ & $+5.00$ & $+2.93$ \\
ReClor & $K{=}8-K{=}0$ & $\mathbf{+9.40}$ & [6.47, 12.53] & $-5.27$ & $+9.27$ & $+5.40$ \\
\midrule
\multicolumn{7}{l}{\textit{B. Randomized inclusion of failures first recovered in samples 5--8, main recipe}} \\
LogiQA 2.0 & $50\%-0\%$ & $+0.80$ & [$-0.90$, 2.61] & $-0.34$ & $+0.32$ & $+0.83$ \\
LogiQA 2.0 & $100\%-0\%$ & $-0.22$ & [$-1.80$, 1.54] & $-1.63$ & $+0.54$ & $+0.88$ \\
ReClor & $50\%-0\%$ & $-0.60$ & [$-2.87$, 1.73] & $-2.13$ & $+1.07$ & $+0.47$ \\
ReClor & $100\%-0\%$ & $+0.20$ & [$-3.53$, 3.93] & $-1.93$ & $+1.33$ & $+0.80$ \\
\midrule
\multicolumn{7}{l}{\textit{C. Failures selected with gold labels, main recipe}} \\
LogiQA 2.0 & gold $-$ GapFT & $\mathbf{-9.49}$ & [$-11.29$, $-7.68$] & $-11.34$ & $-1.07$ & $+2.93$ \\
ReClor & gold $-$ GapFT & $\mathbf{-6.87}$ & [$-12.20$, $-1.40$] & $-6.13$ & $-1.60$ & $+0.87$ \\
\bottomrule
\end{tabular*}
\end{table*}

\paragraph{Replacing replay with recovered failures raises accuracy.}
At a fixed budget, training on failures that search recovers instead of
replaying solved problems raises Pass@1 on both tasks, and the failures the
first sample already recovers, about a quarter of the slots, deliver three
fifths to four fifths of the gain (Table~\ref{tab:channels}, panel A;
Figure~\ref{fig:mechanism_controls}).  The
gain enters through the $S$ and $U$ terms: the trained failures are corrected
and the correction transfers to failures search missed.  Its price is
regression on problems the source already solves, which stays small while most
slots still replay them and grows as replay leaves the budget.  Because $N$ and
$J$ are fixed, the gain comes from which problems are covered, not from more
data or more updates.

\paragraph{Failures found only by deep search trade correction for regression.}
The randomized control asks whether the hardest recovered failures, the quarter
first found in samples five through eight, carry extra value.  At this budget,
including all of them rather than none yields no resolved Pass@1 gain on
either task, with intervals that include zero (panel B;
Figure~\ref{fig:mechanism_controls}, right).  The decomposition
shows why.  The included failures are corrected, but each takes a slot that
would otherwise replay a solved problem, and the added regression cancels the
added correction.  Together with panel A, this places most of the value of
search in its first few samples: at a budget without separate replay, deeper
search adds failures whose correction is paid for by the replay they displace.

\paragraph{Search reachability is itself a selection signal.}
Given gold labels, one could fill the budget with any deployed failure.  At the
same targets, exposures, and updates, doing so lowers Pass@1 on both tasks and
in every seed (Table~\ref{tab:channels}, panel C).  Gold selection does correct
more of the failures that search missed, but it pays far more in regression on
problems the source already solved.  A plausible reason is that a failure no
sample reaches has a target the source assigns little probability to, so
fitting many such targets pulls the policy away from its current decisions,
including correct ones; label noise concentrated in hard failures would act in
the same direction.  Restricting a small budget to $S_K$ therefore does more
than stand in for missing labels: it selects reachable failures with less regression than gold-label selection.  The effect depends on the recipe.  In
the older recipe with a large replayed core, gold-selected failures are not
worse than search, and a ReClor diagnostic that adds the core and more updates to both recipes
closes the gap (Appendix~\ref{app:full_pool_results}); reachability matters
most when the budget holds no replay.

\paragraph{The gain is tied to specific failures, not answer frequencies.}
Answer reshuffling substantially lowers the observed $S_8$ correction, showing
that answer frequencies alone do not explain the effect.  Across four
independent source sample sets, the combined correction remains fixed even
though items move between $S_K$ and $U_K$, so the aggregate effect is stable
while the finer $S/U$ attribution has sampling uncertainty (Appendix
Tables~\ref{tab:scope} and~\ref{tab:stability}).  Together, the controls
indicate that GapFT gains by covering distinct deployed failures that search
can reach.

\FloatBarrier

\section{When GapFT Helps: The Limit of Transferable Failures}
\label{sec:boundaries}
\label{sec:exp_support}
\label{sec:exp_math}
\label{sec:exp_code}

The contribution $p_S\kappa_S$ depends on both the failures search exposes and
their correction after training. GapFT corrects between two fifths and two
thirds of recovered audit failures for all three models, but $S_8$ contains
about a quarter of Llama's audit set, under a tenth of Mistral's, and only
about 2\% of Qwen's (Appendix Table~\ref{tab:main_decomposition}). This share
limits the gain over the source. It need not remove the advantage over uniform
RFT: Mistral benefits from selection because a uniform draw includes few
recovered failures. Qwen leaves little room for either recipe because its
single response already solves nearly every problem that search can reach.

\subsection{Few Transferable Failures Limit the Gain}
\label{sec:math_failure_support}

\begin{table}[H]
\centering
\small
\setlength{\tabcolsep}{4.2pt}
\caption{MATH Results with Matched Training Exposures}
\label{tab:math}
\begin{tabular*}{\textwidth}{@{\extracolsep{\fill}}lrrrrr}
\toprule
Condition & Post Pass@1 (\%) & $\Delta$ vs source (pp) & $\kappa_S$ & $\kappa_U$ & $\rho_G$ \\
\midrule
deployed successes & 63.39 & $+0.56$ & 41.35 & 3.08 & 11.49 \\
search-recovered failures & 66.11 & $+3.28$ & 40.38 & 4.48 & 7.34 \\
uniform RFT & 64.11 & $+1.28$ & 33.65 & 5.60 & 9.02 \\
positive SFT & 66.17 & $+3.33$ & 41.35 & 5.04 & 7.69 \\
verified correction & 64.83 & $+2.00$ & 34.62 & 2.52 & 7.16 \\
188 distinct problems, 4 exposures each & 65.78 & $+2.94$ & 40.06 & 3.92 & 7.60 \\
94 distinct problems, 8 exposures each & 65.17 & $+2.33$ & 40.38 & 2.24 & 8.13 \\
\bottomrule
\end{tabular*}

\end{table}

On the independent 600-problem MATH audit with $K=8$, Qwen2.5-7B trained on
search-recovered failures reaches $66.11\%$ Pass@1, $3.28$ points above the
source and $2.00$ points above matched uniform RFT (Table~\ref{tab:math});
with three seeds, the hierarchical interval for the RFT contrast includes
zero. The gain arises where GapFT directs the budget: relative to uniform RFT,
$S_K$ correction is $40.38\%$ vs.\ $33.65\%$ and $G_K$ regression is
$7.34\%$ vs.\ $9.02\%$. $U_K$ correction is slightly lower ($4.48\%$ vs.\
$5.60\%$), consistent with training on failures that source search already reaches.
The size of the gain on MATH is therefore bounded by how many recoverable
failures search exposes, and regression on previously solved problems still
offsets part of it.

\subsection{Favoring Coverage over Repetition}
\label{sec:coverage_repetition}

At the same number of training exposures, spreading the budget over 188
distinct recoverable failures, four exposures each, gains $2.94$ points over
the source, against $2.33$ points for 94 problems with eight exposures each
(Table~\ref{tab:math}). The $0.61$-point difference has an interval that
includes zero, but its composition favors coverage: $S_K$ correction is nearly
unchanged, while the broader set raises $U_K$ correction from $2.24\%$ to
$3.92\%$ and lowers $G_K$ regression from $8.13\%$ to $7.60\%$. These estimates
favor spending a fixed budget on more distinct recoverable failures rather than
repeating the same ones. Adapter scaling at deployment
(Appendix~\ref{app:scaling_boundary}) only rescales an update already trained
on these problems, so it cannot add failures that source search missed.

\section{Conclusion}
\label{sec:conclusion}

Turning search gains into single-response accuracy requires selecting which
verified responses are worth learning from. GapFT targets the gap between
Pass@$K$ and Pass@1 by selecting problems where the source model can generate
a correct response within $K$ samples but fails in its direct answer.
GapFT fills the fixed training budget with verified $S_K$ responses without
changing the objective; we measure gains from corrected failures against
losses on previously solved problems.  On two
logic benchmarks with three model families, GapFT is above uniform RFT at a
matched budget and recovers most of the full verified pool's gain with a
fraction of its problems.  Further analyses show that the gain comes from
covering distinct failures that search can reach, that failures it cannot
reach cost regressions when the budget holds no replay, and that the gain is
bounded by how many transferable failures search exposes. These results make the gap between search and direct
answering a criterion for allocating verified training data.
\normalsize
\subsection*{AI use statement}
This study uses AI tools to write parts of the code, refine the idea, and adjust the LaTeX formatting.

\section*{Reproducibility Statement}

Appendix~\ref{app:implementation} details the models, data splits, search and
decoding settings, training hyperparameters, budgets, random seeds, and
statistical procedures. Appendix~\ref{app:extended} reports additional
controls and experimental results.
Our code is available at \url{https://github.com/Xuanxuana1/GapFT}.

\bibliography{references}
\bibliographystyle{iclr2027_conference}

\clearpage
\appendix
\raggedbottom

\section{Method Details}
\label{app:implementation}

Tables~\ref{tab:logic_implementation}--\ref{tab:code_implementation}
list the settings needed to reproduce the logic, MATH, and code experiments.
In all three, cross-entropy is computed only on the verified response tokens,
and LoRA \citep{hu2022lora} is applied to the attention projections
\texttt{q\_proj}, \texttt{k\_proj}, \texttt{v\_proj}, and \texttt{o\_proj},
and the MLP projections \texttt{gate\_proj}, \texttt{up\_proj}, and
\texttt{down\_proj}.  All final audits use deterministic decoding.

\paragraph{Logic run families.}
The main comparison selects $n=|\mathcal{E}_{S,8}|$ distinct problems per
condition, retains one response per problem, and repeats each fixed pair
eight times.  Thus $K=8$ is the number of search candidates per problem,
whereas $R=8$ is the number of training exposures per selected pair.  All four
conditions of Table~\ref{tab:main} use training seeds 13, 21, and 73, and the
training prompt reproduces the two start tokens the generator emits.  The
three controls of Section~\ref{sec:mechanism} use the same Llama recipe,
seeds, and $(n,N,J)$.  The search-depth and randomized controls keep GapFT's
response for every included failure and fill the remaining slots with the
first draws of the solved-replay recipe, so each condition differs from GapFT
and from solved replay only in which problems occupy the budget; the
randomized draw of late failures uses a seed fixed per training seed.  On
LogiQA, $K=1$ keeps 831 and the $0\%$ arm 2{,}324 of the 3{,}190 recovered
failures; on ReClor, 241 and 653 of 913.
Appendix~\ref{app:full_pool_results} also reports older Llama controls with
training seeds 13, 21, and 42 and a single start token; we rescore their
frozen outputs on the same audit problems but compare them only with each
other.  Their cumulative
$K=0/1/8$ control retains $\mathcal{E}_G$ and replaces repeated evidence
within $|\mathcal{E}_G|+|\mathcal{E}_{S,8}|$ slots.  Its oracle control
retains the same $\mathcal{E}_G$ and replaces search-selected failures with
the same number of gold-label failures, sampled without replacement within
gold-answer classes.  It assigns the search condition's answer-only target
strings to selected problems with matching labels and repeats the complete
recipe eight times, preserving target string counts and training exposures.
For nested search budgets, the first $K$ samples are reused as a prefix and
the audit partition is rebuilt at each $K$.  Table~\ref{tab:logic_budgets}
separates recipe slots from total training exposures for these older controls.
The scope matrix uses separately completed
uniform RFT checkpoints and selects their adapter scales on disjoint
calibration data.  Absolute post-training accuracies are therefore compared
only within, not across, these protocols.  The scope matrix, the stability
check, the online control, and the full-pool comparison of
Appendix~\ref{app:extended} also predate the main protocol and use the full
1{,}572-problem LogiQA test set, which includes 205 problems whose prompt also
appears in the training split; we keep them as records of that protocol.

\paragraph{Reproducibility.}
Fixed seeds do not make training bitwise reproducible on our hardware.
Retraining one gold-label seed with byte-identical data and settings moves
mean accuracy by $0.07$ points on LogiQA and $1.4$ points on ReClor, but
changes correctness on 331 of 1{,}367 and 71 of 500 audit problems.  We
therefore interpret differences through intervals that resample training
seeds rather than through single runs.

\begin{table}[!htbp]
\caption{Implementation Details for Logic Experiments}
\label{tab:logic_implementation}
\centering
\small
\setlength{\tabcolsep}{5pt}
\begin{tabular}{p{0.29\linewidth}p{0.61\linewidth}}
\toprule
Setting & Value\\
\midrule
Models & Llama-3.1-8B-Instruct, Qwen2.5-7B-Instruct, and Mistral-7B-Instruct-v0.3 in the main comparison; Llama-3.1-8B-Instruct in the controls\\
Search & $K=8$ for the main comparison; $K\in\{0,1,8\}$ for the search-depth control; temperature $0.7$, top-$p=0.95$, at most 64 new tokens; shortest verified response selected, with response index as tie-breaker\\
Verifier / deployment & Exact option matching, on the whole response for Llama and Qwen on LogiQA~2.0 and on the first explicit answer otherwise / greedy choice decoding\\
Update & RFT with loss on response tokens only; AdamW; bfloat16; one epoch over the recipe repeated eight times\\
LoRA & Rank 8, alpha 16, dropout 0.05; seven attention and MLP projection modules listed above\\
Optimization & Learning rate $5\times10^{-5}$; batch size 2; gradient accumulation 8; weight decay 0.01; no scheduler; gradient norm clipped to 1.0; maximum sequence length 2048\\
Matched update budget & $\lceil n/2\rceil$ updates per condition, listed in Table~\ref{tab:logic_budgets}\\
Randomness & Training seeds 13, 21, and 73 in the main comparison and the controls of Section~\ref{sec:mechanism}; 13, 21, and 42 in the older shared-core controls\\
Statistics & Intervals use 2,000 bootstrap replicates that resample training seeds and paired audit problems; inclusion tests and controls use 2,000 permutations; scope matrix answer controls use 10,000 permutations\\
\bottomrule
\end{tabular}
\end{table}

\begin{table}[!htbp]
\centering
\footnotesize
\setlength{\tabcolsep}{4pt}
\caption{Training Budgets of the Logic Experiments}
\label{tab:logic_budgets}
\begin{tabular}{llrrrr}
\toprule
Comparison & Task & $n$ & $N=8n$ & $L$ & $J$ \\
\midrule
Main, Llama & LogiQA 2.0 & 3{,}190 & 25{,}520 & 6{,}049{,}464 & 1{,}595 \\
& ReClor & 913 & 7{,}304 & 1{,}793{,}896 & 457 \\
Main, Qwen & LogiQA 2.0 & 275 & 2{,}200 & 512{,}408 & 138 \\
& ReClor & 63 & 504 & 120{,}064 & 32 \\
Main, Mistral & LogiQA 2.0 & 887 & 7{,}096 & 1{,}901{,}832 & 444 \\
& ReClor & 252 & 2{,}016 & 555{,}896 & 126 \\
\midrule
\multicolumn{6}{l}{\textit{Older shared-core controls, Llama}} \\
Cumulative search & LogiQA 2.0 & 9{,}575 & 76{,}600 & 18{,}049{,}528 & 4{,}788 \\
& ReClor & 3{,}049 & 24{,}392 & 5{,}892{,}872 & 1{,}525 \\
Gold-label control & LogiQA 2.0 & 9{,}575 & 76{,}600 & 17{,}989{,}088 & 4{,}788 \\
& ReClor & 3{,}049 & 24{,}392 & 5{,}864{,}368 & 1{,}525 \\
\bottomrule
\end{tabular}
\end{table}

Here $n$ counts example slots before repetition, $N$ counts total training
exposures, $L$ counts processed input and response tokens, and $J$ counts
optimizer steps.  The main rows list GapFT's processed tokens.  The main
comparison matches $n$, $N$, and $J$ but not tokens: recovered failures have
longer prompts, so GapFT processes 0.4--5.0\% more tokens than uniform RFT or
solved replay.  The cumulative-search rows give processed-token
matching targets; replay replacement can introduce small token-count
discrepancies.  Gold-label selection matches response tokens exactly but
processes $0.33\%$ fewer total tokens on LogiQA and $0.48\%$ fewer on ReClor
than the cumulative $K=8$ control.  All rows use $R=8$ and learning rate
$5\times10^{-5}$.  The all-verified reference in Table~\ref{tab:main} trains
on all $|\mathcal{E}_G|+|\mathcal{E}_{S,8}|$ verified problems in the training
candidate pool with $\lceil n/2\rceil$ updates.

\begin{table}[!htbp]
\caption{MATH Matched-Diversity Control}
\label{tab:math_implementation}
\centering
\small
\setlength{\tabcolsep}{5pt}
\begin{tabular}{p{0.29\linewidth}p{0.61\linewidth}}
\toprule
Setting & Value\\
\midrule
Model & Qwen2.5-7B-Instruct\\
Evidence & 188 distinct failures recovered by search or deployed successes, each repeated four times; subject and difficulty strata differ by at most one problem\\
Update & 752 steps using sequence-normalized cross-entropy on response tokens only; bfloat16; AdamW; constant learning rate $10^{-6}$; weight decay 0.01; batch size 1; maximum sequence length 2048\\
Budget matching & 752 training examples and model positions; 270,912 versus 269,980 supervised response tokens, below the prespecified 2\% tolerance\\
LoRA & Rank 8, alpha 16, dropout 0.05; the seven attention and MLP projection modules listed above\\
Audit & 600 disjoint MATH problems~\citep{hendrycks2021math}; deterministic decoding with at most 1,024 new tokens\\
Randomness / statistics & Training seeds 13, 21, and 42; 10,000 hierarchical bootstrap replicates\\
\bottomrule
\end{tabular}
\end{table}

\begin{table}[!htbp]
\caption{LiveCodeBench Allocation and Replication~\citep{jain2024livecodebench}}
\label{tab:code_implementation}
\centering
\small
\setlength{\tabcolsep}{5pt}
\begin{tabular}{p{0.29\linewidth}p{0.61\linewidth}}
\toprule
Setting & Value\\
\midrule
Models & Qwen2.5-Coder-7B-Instruct~\citep{hui2024qwen25coder} and DeepSeek-Coder-6.7B-Instruct~\citep{guo2024deepseekcoder}\\
Search & $K=16$ independent source episodes; temperature $0.7$, top-$p=0.95$; at most three attempts and 1,536 new tokens per attempt; shortest successful response from the final turn selected, with sample index and seed as tie-breakers\\
Verifier / deployment & All official public and hidden tests / one episode with at most three internal attempts and public test feedback\\
Update & RFT with loss on response tokens only; AdamW; bfloat16; one epoch; gradient checkpointing\\
LoRA & Rank 16, alpha 32, dropout 0.05; the same seven projection modules as the logic experiments\\
Optimization & Learning rate $5\times10^{-5}$; one sequence per microbatch; training examples distributed across exactly 64 update groups; weight decay 0.01; no scheduler; gradient norm clipped to 1.0; maximum sequence length 8192\\
Matched update budget & 64 optimizer updates; matched verified response tokens and identical $G$ replay within each allocation comparison\\
Randomness & Selection and training seeds 13, 21, and 42\\
Inference & Each deterministic episode is re-executed three times to audit environment repeatability; the item label is the majority verifier outcome, and repeats do not add sampled model responses; 20,000 paired bootstrap replicates\\
\bottomrule
\end{tabular}
\end{table}
\FloatBarrier

\section{Extended Experimental Results}
\label{app:extended}

\subsection{Full Logic Scope Matrix}
\label{app:scope_matrix}

Table~\ref{tab:scope} reports both search budgets for the scope matrix with
two tasks and two models.  Conditional rates with fewer than 30 $S_K$ items are
marked because their intervals are unstable.  The compact $K=8$ summary in the
main text reports the main denominators; this table gives the complete
accounting from the source to the post-training policy.

\begin{table}[!htbp]
\centering
\small
\setlength{\tabcolsep}{3.6pt}
\caption{Logic Scope Matrix and Answer-Redistribution Control}
\label{tab:scope}
\resizebox{\textwidth}{!}{%
\begin{tabular}{llrlrrrrrrr}
\toprule
& & & & & \multicolumn{3}{c}{observed rates (\%)} & & \multicolumn{2}{c}{redistribution null (\%)} \\
\cmidrule(lr){6-8}\cmidrule(lr){10-11}
Dataset & Model & $K$ & $G/S/U$ & $p_S$ (\%) & $\kappa_S$ & $\kappa_U$ & $\rho_G$ & Net gain (pp) & $S$ & $U$ \\
\midrule
LogiQA 2.0 & Llama-3.1-8B & 4 & 770 / 306 / 496 & 19.47 & 47.17 & 20.50 & 8.23 & $+11.62$ & -- & -- \\
LogiQA 2.0 & Llama-3.1-8B & 8 & 770 / 416 / 386 & 26.46 & 50.72 & 16.93 & 6.88 & $+14.21$ & 21.90 & 11.11 \\
LogiQA 2.0 & Qwen2.5-7B & 4 & 1{,}050 / 28 / 494 & 1.78 & 51.19 & 8.37 & 2.79 & $+1.68$ & -- & -- \\
LogiQA 2.0 & Qwen2.5-7B & 8 & 1{,}050 / 37 / 485 & 2.35 & 54.95 & 7.56 & 2.73 & $+1.80$ & 27.83 & 4.92 \\
ReClor & Llama-3.1-8B & 4 & 277 / 96 / 127 & 19.20 & 46.88 & 38.58 & 6.38 & $+15.27$ & -- & -- \\
ReClor & Llama-3.1-8B & 8 & 277 / 130 / 93 & 26.00 & 55.64 & 38.71 & 8.66 & $+16.87$ & 23.86 & 21.16 \\
ReClor & Qwen2.5-7B & 4 & 361 / 7 / 132 & 1.40 & 71.43 & 4.04 & 1.20 & $+1.20$ & -- & -- \\
ReClor & Qwen2.5-7B & 8 & 361 / 11 / 128 & 2.20 & 57.58 & 3.65 & 1.11 & $+1.40$ & 24.24 & 2.28 \\
\bottomrule
\end{tabular}}
\end{table}

For example, at LogiQA/$K=4$, Llama's $S$ term within observed support contributes $9.18$
points and Qwen's contributes only $0.91$ points.  The difference primarily
comes from the $306$ versus $28$ audit items in $S_4$, rather than from a low
Qwen conditional correction rate.  From $K=4$ to $K=8$, the number of
$S$ items in the training candidate pool changes from $2{,}364$ to $3{,}249$ for LogiQA/Llama,
$655$ to $916$ for ReClor/Llama, $209$ to $279$ for LogiQA/Qwen, and $50$ to
$63$ for ReClor/Qwen.  These counts make newly available verified evidence,
rather than nominal $K$ alone, the appropriate basis for reading the matrix.

\subsection{Comparison with Closely Related Methods}
\label{app:related_comparison}

Table~\ref{tab:related_comparison} contrasts GapFT with the methods most
likely to be compared against it, along the dimensions of research question,
assumption about verified evidence, mechanism, training regime, and the
difference from GapFT.

\begin{table}[!htbp]
\centering
\footnotesize
\setlength{\tabcolsep}{3pt}
\caption{Comparison with Closely Related Methods}
\label{tab:related_comparison}
\renewcommand{\arraystretch}{1.15}
\begin{tabular}{>{\raggedright\arraybackslash}p{1.8cm}>{\raggedright\arraybackslash}p{2.1cm}>{\raggedright\arraybackslash}p{2.4cm}>{\raggedright\arraybackslash}p{2.4cm}>{\raggedright\arraybackslash}p{1.1cm}>{\raggedright\arraybackslash}p{2.8cm}}
\toprule
Method & Question & Assumption on verified evidence & Mechanism & Regime & Difference from GapFT \\
\midrule
RFT \citep{shao2024deepseekmath} & Improve reasoning from self-generated data & All verified responses are useful & Sample, keep verifier-accepted responses, fine-tune & Offline & Ignores state; budget not matched across evidence sets \\
Distinct paths \citep{yuan2023scaling} & What drives RFT gains & Distinct correct paths matter more than samples & Augment with diverse verified paths & Offline & Diversity of responses, not state of the problem \\
DART-Math \citep{tong2024dartmath} & Correct the easy-query bias of rejection sampling & Difficult queries are under-covered & Give more synthesis trials to queries with a high fail rate under a separate synthesis model & Offline & Selects by query difficulty for the synthesizer, not the deployed state of the trained policy; budget not matched \\
DAPO dynamic sampling \citep{yu2025dapo} & Keep gradient signal in group RL & Prompts with uniform group outcome carry no signal & Drop all-correct and all-incorrect prompts online & Online & Filters on current-policy reward variance, which Appendix~\ref{app:online_control} shows is not the source state \\
BOND \citep{sessa2025bond} & Amortize Best-of-$N$ & The Best-of-$N$ distribution is the target & Distill Best-of-$N$ via Jeffreys divergence & Online & Targets a distribution, not the failure set; no state distinction \\
SCoRe \citep{kumar2024score} & Self-correction at test time & Off-policy correction traces mismatch the policy & Two-stage on-policy RL with reward shaping & Online & Repairs mismatch by RL; GapFT selects evidence by state offline \\
Pass@$k$ optimization \citep{2026passk_degrades_pass1} & Why Pass@$k$ training hurts Pass@1 & Objective rewards any correct sample & Analyzes prompt interference in the objective & Online & Objective-level mismatch; GapFT addresses evidence-level mismatch and measures $\rho_G$ \\
GapFT (ours) & Value of a verified response under single-sample decoding & Value depends on state & Fill a matched budget with $S_K$ only; audit by exact decomposition & Offline & -- \\
\bottomrule
\end{tabular}
\end{table}

\subsection{Evidence Replacement Controls}
\label{app:full_pool_results}

Table~\ref{tab:main_full_pool} reports the older cumulative search
and gold-label controls.  They retain $\mathcal{E}_G$, use
$n=|\mathcal{E}_G|+|\mathcal{E}_{S,8}|$ slots, and replace replay with
source-failure evidence.  Search budgets match processed tokens; the
gold-label comparison matches response tokens as detailed in
Table~\ref{tab:logic_budgets}.  LogiQA entries are rescored on the 1{,}367
deduplicated audit problems.

With the replayed core, gold-selected failures are not worse than search:
the gold-label recipe minus the search recipe is $+1.05$ $[-0.71, 2.61]$ points
on LogiQA and $+5.27$ $[2.93, 7.53]$ on ReClor, the reverse of the ordering in
Table~\ref{tab:channels}, panel C.  Besides the core, these runs differ from
the main recipe in training seeds, the number of updates, and the start tokens
of the training prompt.  A two-seed ReClor diagnostic adds the same 2{,}133-problem
core, with eight exposures each, to both the gold-label and the GapFT recipe of
the main protocol, raising updates from 457 to 1{,}523.  Gold selection then
rises from 63.70\% to 71.90\% and GapFT from 70.60\% to 72.40\%, a difference of
$-0.50$ $[-3.10, 2.00]$ points, and the recovered accuracy comes almost entirely
from fewer regressions on $G_K$.  This diagnostic does not separate the core
data from the extra updates, and it does not reproduce the full old ordering, so
we read it only as evidence that the disadvantage of gold selection depends on
whether the budget also replays deployed successes.

\begin{table}[H]
\centering
\small
\setlength{\tabcolsep}{4.5pt}
\caption{Pass@1 of the Evidence Controls with a Shared Core}
\label{tab:main_full_pool}
\resizebox{\textwidth}{!}{%
\begin{tabular}{lrlrl}
\toprule
& \multicolumn{2}{c}{LogiQA 2.0} & \multicolumn{2}{c}{ReClor} \\
\cmidrule(lr){2-3}\cmidrule(lr){4-5}
Training condition & Pass@1 (\%) & $\Delta$ vs replay (pp) & Pass@1 (\%) & $\Delta$ vs replay (pp) \\
\midrule
no post-training (source) & 50.33 & -- & 55.40 & -- \\
solved replay, $K{=}0$ & 54.18 & 0 (ref.) & 59.27 & 0 (ref.) \\
search evidence, $K{=}1$ & 56.91 & $\mathbf{+2.73}$ [1.51, 3.93] & 62.40 & $\mathbf{+3.13}$ [0.47, 5.60] \\
search evidence, $K{=}8$ & 62.33 & $\mathbf{+8.14}$ [6.78, 9.61] & 67.93 & $\mathbf{+8.67}$ [6.27, 10.93] \\
failures selected with gold labels & 63.37 & $\mathbf{+9.19}$ [7.14, 11.09] & 73.20 & $\mathbf{+13.93}$ [11.33, 16.67] \\
\bottomrule
\end{tabular}}
\end{table}

Table~\ref{tab:main_decomposition} decomposes
the main comparison of Table~\ref{tab:main}, and Table~\ref{tab:amortization}
places each of its conditions on the source's Pass@$K$ curve.

\begin{table}[H]
\centering
\footnotesize
\setlength{\tabcolsep}{3pt}
\caption{Decomposition of the Main Comparison}
\label{tab:main_decomposition}
\begin{tabular*}{\textwidth}{@{\extracolsep{\fill}}lllrrrrrrr}
\toprule
& & & & \multicolumn{3}{c}{Rates (\%)} & \multicolumn{3}{c}{Term difference vs uniform (pp)} \\
\cmidrule(lr){5-7}\cmidrule(lr){8-10}
Model & Task & $p_S$ (\%) & Condition & $\kappa_S$ & $\kappa_U$ & $\rho_G$ & $S$ & $U$ & $G$ \\
\midrule
Llama-3.1-8B & LogiQA 2.0 & 26.6 & replay & 30.7 & 4.4 & 7.3 & $-4.22$ & $-2.88$ & $+2.61$ \\
& & & uniform & 46.5 & 16.9 & 12.5 & ref. & ref. & ref. \\
& & & GapFT & 54.7 & 31.0 & 18.8 & $+2.17$ & $+3.24$ & $-3.15$ \\
& & & all verified & 56.1 & 26.7 & 11.0 & $+2.56$ & $+2.24$ & $+0.76$ \\
\midrule
Llama-3.1-8B & ReClor & 26.0 & replay & 27.4 & 15.8 & 7.6 & $-5.73$ & $-3.60$ & $+1.47$ \\
& & & uniform & 49.5 & 35.1 & 10.2 & ref. & ref. & ref. \\
& & & GapFT & 63.1 & 44.8 & 17.1 & $+3.53$ & $+1.80$ & $-3.80$ \\
& & & all verified & 58.7 & 47.3 & 10.2 & $+2.40$ & $+2.27$ & $0.00$ \\
\midrule
Qwen2.5-7B & LogiQA 2.0 & 2.1 & replay & 29.9 & 1.3 & 1.6 & $-0.07$ & $-0.05$ & $+0.41$ \\
& & & uniform & 33.3 & 1.5 & 2.2 & ref. & ref. & ref. \\
& & & GapFT & 44.8 & 6.1 & 3.6 & $+0.24$ & $+1.34$ & $-0.98$ \\
& & & all verified & 56.3 & 8.1 & 3.0 & $+0.49$ & $+1.93$ & $-0.56$ \\
\midrule
Qwen2.5-7B & ReClor & 2.2 & replay & 39.4 & 1.6 & 2.0 & $+0.13$ & $-0.80$ & $+1.13$ \\
& & & uniform & 33.3 & 4.7 & 3.6 & ref. & ref. & ref. \\
& & & GapFT & 42.4 & 4.9 & 2.7 & $+0.20$ & $+0.07$ & $+0.67$ \\
& & & all verified & 48.5 & 5.2 & 1.9 & $+0.33$ & $+0.13$ & $+1.20$ \\
\midrule
Mistral-7B-v0.3 & LogiQA 2.0 & 7.8 & replay & 36.8 & 5.4 & 5.9 & $-0.02$ & $-0.61$ & $+0.80$ \\
& & & uniform & 37.1 & 6.7 & 7.6 & ref. & ref. & ref. \\
& & & GapFT & 45.9 & 10.3 & 7.6 & $+0.68$ & $+1.61$ & $0.00$ \\
& & & all verified & 43.1 & 11.8 & 5.7 & $+0.46$ & $+2.29$ & $+0.88$ \\
\midrule
Mistral-7B-v0.3 & ReClor & 6.8 & replay & 52.9 & 10.2 & 7.1 & $+0.27$ & $+0.40$ & $+0.27$ \\
& & & uniform & 49.0 & 9.3 & 7.7 & ref. & ref. & ref. \\
& & & GapFT & 43.1 & 14.6 & 7.9 & $-0.40$ & $+2.33$ & $-0.13$ \\
& & & all verified & 57.8 & 17.1 & 6.4 & $+0.60$ & $+3.47$ & $+0.60$ \\
\bottomrule
\end{tabular*}
\end{table}

The term differences are the three summands of
Equation~\ref{eq:decomposition} minus those of uniform RFT, averaged over
seeds, so each row sums to its Pass@1 difference from uniform RFT in
Table~\ref{tab:main}; a negative $G$ entry means more regression.  With 29 and
11 audit problems in $S_8$, the Qwen rates $\kappa_S$ are unstable.

\begin{table}[H]
\centering
\small
\setlength{\tabcolsep}{3pt}
\caption{Source Pass@$K$ Reached by One Decode}
\label{tab:amortization}
\begin{tabular}{llrrrrrrr}
\toprule
& & \multicolumn{3}{c}{Source (\%)} & \multicolumn{4}{c}{$K_{\mathrm{eq}}$} \\
\cmidrule(lr){3-5}\cmidrule(lr){6-9}
Model & Task & Greedy & Pass@4 & Pass@8 & Replay & Uniform & GapFT & All \\
\midrule
Llama-3.1-8B & LogiQA 2.0 & 50.33 & 68.40 & 76.81 & 1 & 2 & 2 & 3 \\
& ReClor & 55.40 & 73.32 & 81.40 & 1 & 2 & 3 & 4 \\
Qwen2.5-7B & LogiQA 2.0 & 68.54 & 70.17 & 70.67 & $<1$ & $<1$ & 1 & 3 \\
& ReClor & 72.20 & 73.69 & 74.40 & $<1$ & $<1$ & 1 & 2 \\
Mistral-7B-v0.3 & LogiQA 2.0 & 47.26 & 52.53 & 54.94 & 1 & 1 & 3 & 4 \\
& ReClor & 48.80 & 53.78 & 55.60 & 3 & 2 & 4 & 8 \\
\bottomrule
\end{tabular}
\end{table}

$K_{\mathrm{eq}}$ is the largest $K$ whose source Pass@$K$, estimated without
bias from the eight audit samples \citep{chen2021codex}, does not exceed the
condition's Pass@1; $<1$ means the condition is below the source's sampled
Pass@1.  Pass@$K$ is an oracle reference that needs search and a verifier at
inference.

\begin{figure}[H]
\centering
\includegraphics[width=\linewidth]{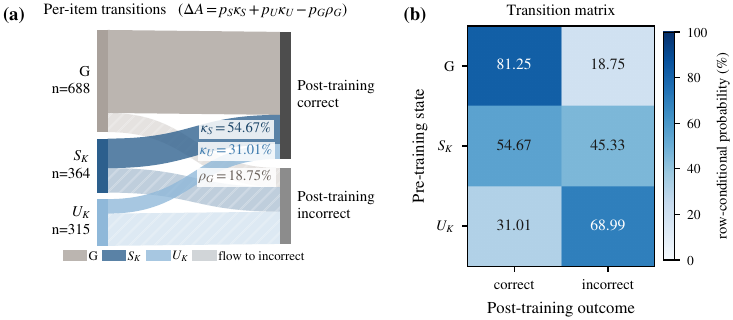}
\caption{\textbf{Outcome flows by source state.} (a) Where the audit
problems of each source state end up after GapFT; (b) the same transitions as
rates within each source state.  Llama-3.1-8B on LogiQA 2.0, averaged over
three training seeds on the deduplicated audit set.}
\label{fig:audit_flow}
\end{figure}

Figure~\ref{fig:audit_flow} illustrates the decomposition flows.

\begin{figure}[H]
\centering
\includegraphics[width=\textwidth]{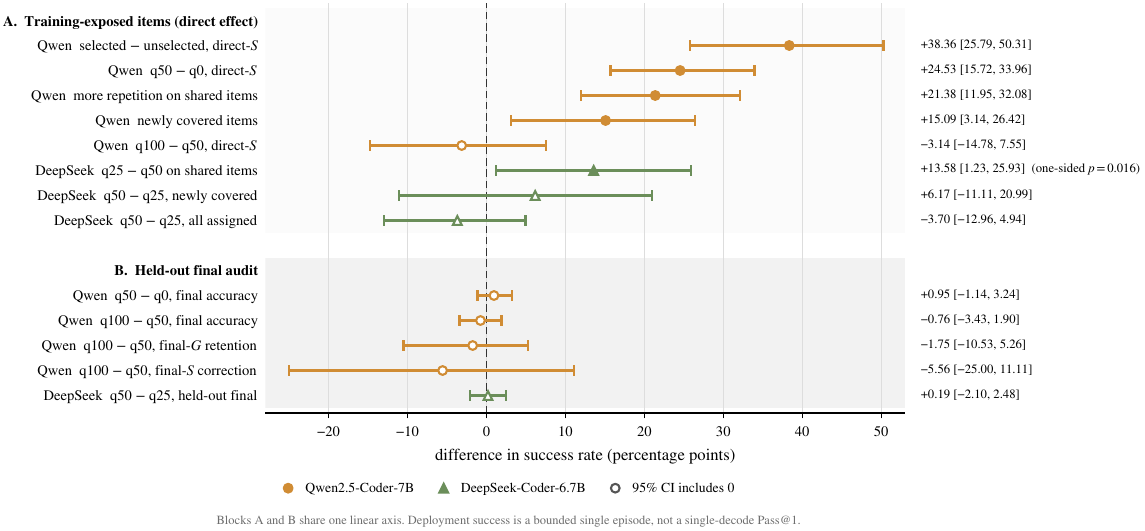}
\caption{Code Coverage and Transfer Effects}
\label{fig:code_boundary}
\end{figure}

Figure~\ref{fig:code_boundary} shows the code decomposition of
Section~\ref{sec:exp_code}.
For shared half $A$, new half $B$, and subset success $a_C(r)$ after $r$
exposures, the code allocation contrast is
\begin{equation}
\Delta_{\mathrm{diffuse-concentrated}}
=\tfrac12\underbrace{[a_B(1)-a_B(0)]}_{\text{coverage gain}}
-\tfrac12\underbrace{[a_A(2)-a_A(1)]}_{\text{repetition loss}}.
\label{eq:coverage_repetition}
\end{equation}

\begingroup
\raggedbottom
\subsection{Qualitative Cases: Before-and-After Solutions}
\label{app:before_after}

The cases below show how generated solutions change after training.
They come from an existing Qwen2.5-7B rerun of the MATH recipe in
Table~\ref{tab:math}: 188 search-recovered failures, four exposures per problem,
752 updates, and a constant learning rate of $10^{-6}$ (seed 13).
This rerun used a separate 600-problem held-out set; these are not the
outputs used to compute Table~\ref{tab:math}.
Each pair contains the complete greedy response before training and after
GapFT, with a short task summary. Red marks erroneous steps or answers;
teal marks their correct counterparts. We chose two concise examples of
a correction and a regression; uniform RFT has the same correctness
transitions on these items.

\par\medskip\noindent\fbox{%
\begin{minipage}{\dimexpr\linewidth-2\fboxsep-2\fboxrule\relax}
\small\setlength{\parskip}{3pt}
\textbf{Case 1: Correcting the product and considering both signs.}\par
MATH algebra:249; source state $S_8$.\par
\textit{Task summary.} The roots of $x^2+bx+18=0$ have ratio $2:1$. Find the largest $b$.\par\medskip
\begin{minipage}[t]{0.48\linewidth}\vspace{0pt}
\textbf{Before: source model (wrong)}\par\smallskip
\raggedright Let the roots be $2r$ and $r$. By Vieta's formulas, $(2r + r) = -b$ and $(2r \cdot r) = 18$. {\color{red!65!black}Thus, $3r^2 = 18$, so $r^2 = 6$ and $r = \sqrt{6}$.} Then, $-b = 3r = 3\sqrt{6}$, giving $b = -3\sqrt{6}$. \par\smallskip\noindent Therefore, the largest possible value of $b$ is: {\color{red!65!black}$\boxed{-3\sqrt{6}}$}.\par
\end{minipage}
\hfill%
\begin{minipage}[t]{0.48\linewidth}\vspace{0pt}
\textbf{After: GapFT (correct)}\par\smallskip
\raggedright Let the roots be $2r$ and $r$. By Vieta's formulas, their sum is $-b$ and their product is $18$.\par\noindent {\color{teal!70!black}So, $(2r)(r) = 2r^2 = 18 \Rightarrow r^2 = 9 \Rightarrow r = 3$ or $r = -3$.}\par\noindent For $r = 3$, the roots are $6$ and $3$, giving $-b = 6 + 3 = 9 \Rightarrow b = -9$.\par\noindent {\color{teal!70!black}For $r = -3$, the roots are $-6$ and $-3$, giving $-b = -6 - 3 = -9 \Rightarrow b = 9$.}\par\noindent Therefore, the largest possible value of $b$ is: {\color{teal!70!black}$\boxed{9}$}.\par
\end{minipage}
\par\medskip\textbf{Change.} The product becomes $2r^2$, and the negative-root branch gives the maximum $b=9$.
\end{minipage}}\par\medskip

\par\medskip\noindent\fbox{%
\begin{minipage}{\dimexpr\linewidth-2\fboxsep-2\fboxrule\relax}
\small\setlength{\parskip}{3pt}
\textbf{Case 2: Keeping the calculation but answering the wrong quantity.}\par
MATH number\_theory:202; source state $G$.\par
\textit{Task summary.} Pool 239, 174, and 83 marbles into piles of ten. How many marbles must be removed?\par\medskip
\begin{minipage}[t]{0.48\linewidth}\vspace{0pt}
\textbf{Before: source model (correct)}\par\smallskip
\raggedright 1. Calculate total marbles: \(239 + 174 + 83 = 496\).\par\noindent 2. Determine how many full piles of 10 can be made: \(496 \div 10 = 49\) remainder \(6\).\par\noindent 3. To form complete piles, remove the remainder: \(496 - 6 = 490\).\par\noindent {\color{teal!70!black}4. Therefore, the number of marbles to remove is: \(496 - 490 = 6\).}\par\smallskip\noindent Therefore, the final answer is: {\color{teal!70!black}$\boxed{6}$}.\par
\end{minipage}
\hfill%
\begin{minipage}[t]{0.48\linewidth}\vspace{0pt}
\textbf{After: GapFT (wrong)}\par\smallskip
\raggedright 1. Calculate total marbles: \(239 + 174 + 83 = 496\).\par\noindent 2. Determine how many full piles of 10 can be made: \(496 \div 10 = 49\) remainder \(6\).\par\noindent 3. To form complete piles, remove the remainder: \(496 - 6 = 490\).\par\smallskip\noindent Therefore, the final answer is: {\color{red!65!black}$\boxed{490}$}.\par
\end{minipage}
\par\medskip\textbf{Change.} The final answer switches from the six removed marbles to the 490 that remain.
\end{minipage}}\par\medskip

\endgroup

\subsection{Online Optimization at Matched Update Steps}
\label{app:online_control}

One control replaces the offline objective with online RL while holding the
number of optimizer updates fixed.  On LogiQA~2.0 with the same source model
and the full audit set, GRPO \citep{shao2024deepseekmath} runs for the $1{,}625$
updates of the earlier constrained protocol with group size $8$, LoRA $r=8$, rule-based
reward on the option letter, and three seeds.  One condition uses every
prompt in the training candidate pool; the other masks groups whose rollouts all receive the same
reward, the filtering rule of DAPO \citep{yu2025dapo}, after the rollouts are
generated, so the two conditions differ in which groups reach the gradient and
not in sampling cost.

Two properties of this control limit what it can establish.  First, it fixes
updates rather than generation: each seed produces about $2.8$M completion
tokens against the $6.1$M processed tokens of a supervised condition, with a
$64$-token cap, so the online conditions see less generation than the offline
ones and complete about half an epoch.  Second, the reward uses a verifier that
accepts an option letter followed by its text, whereas the audit protocol of
Section~\ref{sec:exp_setup} accepts the letter alone.  The online conditions
drift toward the longer format their reward permits, which the audit scores as
invalid: $19.3\%$ and $65.2\%$ of their audit responses against $0.13\%$ for
GapFT, and about half of the invalid responses open with the correct
letter.  Table~\ref{tab:online_control} therefore scores every condition under
both rules.

\begin{table}[!htbp]
\centering
\footnotesize
\setlength{\tabcolsep}{3pt}
\caption{Online RL Control at Matched Updates}
\label{tab:online_control}
\begin{tabular}{lrrrr}
\toprule
& \multicolumn{2}{c}{Answer only} & \multicolumn{2}{c}{Permissive} \\
\cmidrule(lr){2-3}\cmidrule(lr){4-5}
Condition & Pass@1 (\%) & $\Delta$ vs GapFT & Pass@1 (\%) & $\Delta$ vs GapFT \\
\midrule
source & 48.98 & -- & 49.36 & -- \\
GapFT & 63.40 & ref. & 63.40 & ref. \\
GRPO, all prompts & 42.56 & $-20.84$ [$-23.1$,\,$-18.5$] & 51.55 & $-11.85$ [$-14.2$,\,$-9.7$] \\
GRPO, flat groups masked & 19.02 & $-44.38$ [$-46.6$,\,$-42.1$] & 47.43 & $-15.97$ [$-18.3$,\,$-13.7$] \\
\bottomrule
\end{tabular}
\end{table}

Both online conditions stay below GapFT under either rule, and the
permissive rule places them near the source model rather than far below it,
so the answer-only gap is mostly format drift and not lost accuracy.  The
masking rule also fails to reproduce evidence selection: deployed successes
are $29.1\%$ of the groups it drops, far from the $67.9\%$ that dropping every
deployed success and every failure search misses would give, because reward
flatness is a property of the current policy at sampling temperature while the
evidence state is a property of the source model at a single sample.  Neither
observation bounds online RL in general: the control covers one algorithm at
one generation budget, and evidence selection acts on which problems enter
training rather than on the objective, so applying it to online prompt
selection is untested.

\subsection{Allocation When the Budget Covers the Pool}
\label{app:full_pool}

At $n=|\mathcal{E}_G|+|\mathcal{E}_{S,8}|$, a separate comparison trains on
every verified problem once per repetition or samples uniformly with
replacement from the same pool.  The latter includes duplicate problems, so
the comparison tests coverage versus repetition at a larger budget.
The paired contrast, all verified minus uniform with replacement, is $+0.91$
[$-0.45$, $2.21$] on LogiQA and $-2.53$ [$-5.13$, $0.27$] on ReClor.
The main comparison instead fixes $n=|\mathcal{E}_{S,8}|$ and selects without
replacement, testing whether source-state selection improves accuracy when
only part of the verified pool can be used.  The all-verified condition remains
a reference with a larger budget in Table~\ref{tab:main}.

\subsection{Stability Across Independent Source Samples}
\label{app:source_sample_stability}

The $S_K/U_K$ distinction depends on a finite sample from the source model.
Table~\ref{tab:stability} repeats that sample four times while holding
greedy source and post-training decisions fixed.

\begin{table}[!htbp]
\centering
\small
\caption{Stability of the Source-State Partition}
\label{tab:stability}
\setlength{\tabcolsep}{4.1pt}
\begin{tabular}{lrr}
\toprule
Quantity & LogiQA 2.0 & ReClor \\
\midrule
\multicolumn{3}{l}{\textit{A. Partition and flows for each item}} \\
Unstable $S/U$ share among items the source gets wrong (\%) & 35.91 & 44.39 \\
Mean $S/U$ agreement across partitions (\%) & 79.90 & 74.89 \\
$S$ Jaccard across partitions (\%) & 67.67 & 63.87 \\
Mean $S$ flow (pp) & 15.07 & 14.87 \\
Mean $U$ flow (pp) & 6.39 & 7.53 \\
Combined correction (pp) & 21.46 & 22.40 \\
Total gain (pp) & 16.35 & 17.47 \\
$G$ regression tax (pp) & 5.11 & 4.93 \\
\midrule
\multicolumn{3}{l}{\textit{B. Range across four partitions at $K=8$ (pp)}} \\
Total gain & 0.00 & 0.00 \\
$G$ regression & 0.00 & 0.00 \\
Combined correction & 0.00 & 0.00 \\
$S$ flow / $U$ flow & 0.30 & 2.73 \\
$\kappa_S$ & 1.37 & 6.55 \\
$\kappa_U$ & 1.24 & 10.73 \\
\bottomrule
\end{tabular}
\end{table}
\FloatBarrier

Source-sample variation reallocates at most $1.38$ LogiQA points and $2.80$ ReClor
points between the $S$ and $U$ terms.  Their combined correction on items the
source gets wrong and the final accuracy are exactly invariant because the underlying decisions
do not change.  Thus, $\kappa_S$ and $\kappa_U$ must be interpreted conditional
on the sampled source partition, while the matched-budget Pass@1 contrasts remain
unchanged.

\section{Additional Protocol Details}

\subsection{Adapter Scaling with a Calibration Constraint}
\label{app:scaling_boundary}

This rule is used for the scope matrix and its answer redistribution and
source sample controls.  It is not used for the search-depth or randomized
inclusion controls.

For a completed LoRA update $\Delta$, deployment at scale $\alpha$ uses
\begin{equation}
\theta(\alpha)=\theta_0+\alpha\Delta,
\qquad \alpha\in\mathcal{A}=\{0,.25,.5,.75,1\}.
\end{equation}
No retraining occurs.  Each checkpoint selects its scale using a calibration
split only.  For every nonzero $\alpha$, let $u_\alpha$ be a one-sided Wilson
upper confidence bound for $\rho_G(\alpha)$ with Bonferroni correction over
the four nonzero comparisons.  A scale is \emph{safe for every smaller scale} when
\begin{equation}
\max_{\beta\in\mathcal{A}:0<\beta\leq\alpha}u_\beta\leq 0.10.
\end{equation}
The rule selects the largest scale that is safe for every smaller scale using calibration data only and
falls back to $\alpha=0$ if no nonzero scale is safe.  Across 24 checkpoints
from eight task--model--budget environments,
17 select $\alpha=1$, six select $\alpha=.75$, and one selects $\alpha=.5$.
The equal-environment mean improves over the source policy by $8.00$ points with
hierarchical 95\% CI $[3.35,12.76]$, but differs from fixed $\alpha=.75$ by
only $+0.38$ points, CI $[-0.17,0.86]$.  It reduces the number of checkpoints
with a $\rho_G>10\%$ point estimate from four to one, but the remaining case
reaches $10.11\%$ on the final split.  Calibration scaling reduces observed
regression but does not satisfy the universal risk criterion.

\end{document}